\documentclass{article}
\usepackage{acra}
\usepackage[T1]{fontenc} 
\usepackage{hyperref} 
\hypersetup{colorlinks=true, allcolors={blue}} 
\usepackage{amsmath, amssymb, amsfonts} 
\usepackage[linesnumbered,ruled,vlined]{algorithm2e} 
\SetCommentSty{mycommfont} 
\usepackage{graphicx} 
\usepackage{subfig} 
\usepackage{booktabs} 
\usepackage{tabularx} 
\newcolumntype{C}{>{\centering\arraybackslash}X} 
\usepackage{caption}
\usepackage{subcaption}

\title{Harnessing Bounded-Support Evolution Strategies for Policy Refinement}

\author{
    Ethan Hirschowitz \\ The University of Sydney, Australia \\ \textit{ehir9923@uni.sydney.edu.au}
    \And Fabio Ramos \\ 
    The University of Sydney, Australia \\ 
    NVIDIA, USA \\
    \textit{fabio.ramos@sydney.edu.au}
}

\begin{document}

\maketitle

\begin{abstract}
Improving competent robot policies with on-policy RL is often hampered by noisy, low-signal gradients.
We revisit Evolution Strategies (ES) as a policy-gradient proxy and localize exploration with bounded, antithetic triangular perturbations, suitable for policy refinement.
We propose Triangular-Distribution ES (TD-ES) which pairs bounded triangular noise with a centered-rank finite-difference estimator to deliver stable, parallelizable, gradient-free updates. 
In a two-stage pipeline --- PPO pretraining followed by TD-ES refinement --- this preserves early sample efficiency while enabling robust late-stage gains. 
Across a suite of robotic manipulation tasks, TD-ES raises success rates by 26.5\% relative to PPO and greatly reduces variance, offering a simple, compute-light path to reliable refinement.
\end{abstract}

\section{Introduction}

Reinforcement learning (RL) equips robots with the ability to acquire complex behaviors directly from interaction, sidestepping laborious analytic modeling and expensive data collection processes \cite{Ibarz2021-dd,Mirchandani2024-sy}. 
While gradient-based deep RL has achieved impressive results in dexterous manipulation, legged locomotion and autonomous driving domains~\cite{OpenAI2018-gz,Tang2024-rk,Wang2023-ct}, its practical deployment remains a challenge~\cite{Henderson2017-ji,Dulac-Arnold2019-cf,Fu2020-ek,Engstrom2020-dk}.

Crucially, popular policy-gradient methods such as PPO~\cite{Schulman2017-qh} approximate gradients from stochastic rollouts using surrogate objectives and clipping. 
In a refinement context, when near competent policies, these estimates become small-magnitude, high-variance and sensitive to hyperparameters, making fine-grained refinement unreliable and costly \cite{Engstrom2020-dk,Wolczyk2024-gr,Kirkpatrick2017-ln}. 
Yet in robotics applications, even modest performance gains can yield considerable practical value \cite{Dulac-Arnold2019-cf}.

\begin{figure}[t]
\centering
\includegraphics[width=\linewidth]{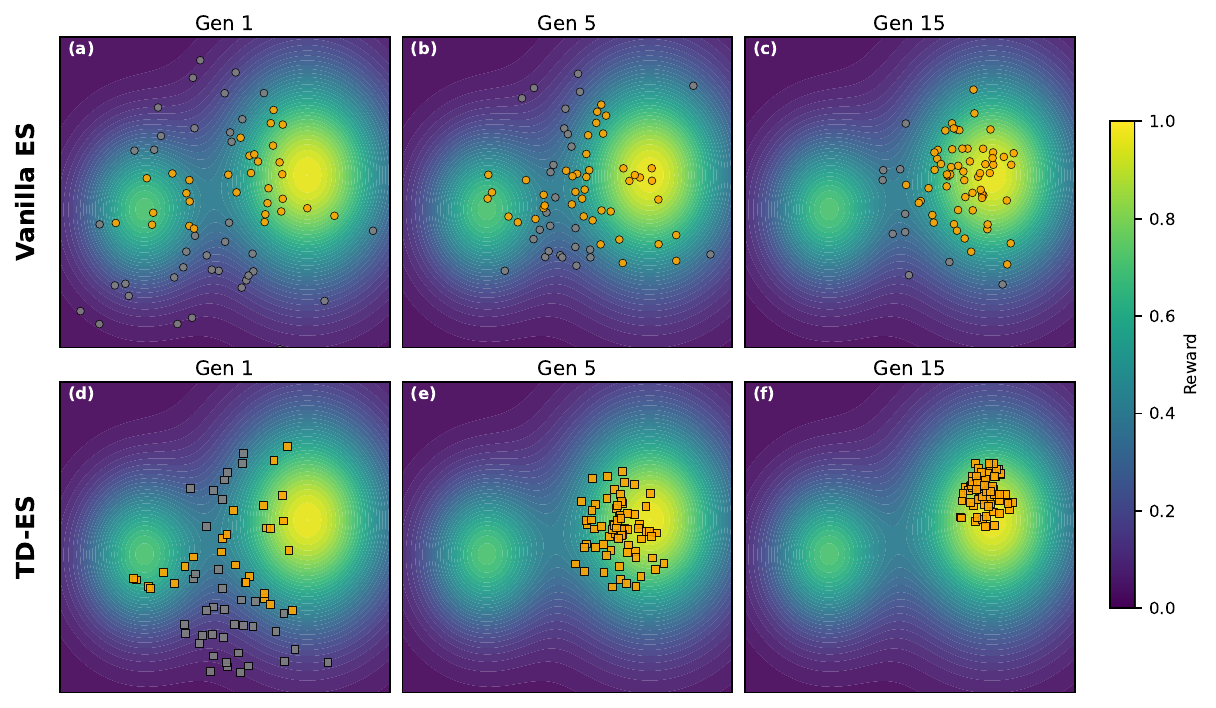}
\caption{Parameter space exploration comparison between Gaussian ES and TD-ES across generations. Both methods search from a PPO checkpoint toward higher-reward regions. Gaussian ES (top row) uses unbounded perturbations that spread widely, while TD-ES (bottom row) employs bounded triangular perturbations for localized exploration. Orange points indicate candidates in high-reward regions, grey points in lower-reward areas. TD-ES achieves more focused exploration with higher sample efficiency in beneficial regions.}
\label{fig:method-overview}
\end{figure}

Evolution Strategies (ES)~\cite{Beyer-firstES,Wierstra2014-kd,Salimans2017-nv} show promise for policy refinement due to their gradient-free nature and robustness to local optima, making them well-suited to refining competent policies where gradient-based methods may struggle with small, noisy signals. 
However, standard ES with unconstrained (e.g. Gaussian) perturbations is poorly matched to refinement: much of the sampling mass lies far from the current parameters, inflating estimator variance and reducing refinement efficiency \cite{Wong2024-ms,Majid2021-rd,Pagliuca2019-di}.

We propose a two-stage refinement pipeline that couples the data efficiency of gradient methods with the stability and parallelism of a gradient-free search. Stage~1 uses PPO to obtain a competent policy and coarse ascent directions. Stage~2 introduces \emph{Triangular-Distribution Evolution Strategies} (TD-ES): an ES variant that samples parameter-space perturbations from a bounded-support triangular distribution, concentrating probability near zero while enforcing a hard radius on updates. This implements a soft trust region in parameter space, reducing estimator variance, mitigating catastrophic regressions, and focusing exploration where the objective is locally smooth, as illustrated in Figure~\ref{fig:method-overview}. \textbf{Specifically, we make the following contributions:}

\begin{enumerate}
    \item We formulate TD-ES, a localized ES refinement method using antithetic, unit-variance triangular perturbations and a centered-rank finite-difference estimator, yielding a simple, compute-light gradient proxy constructed from scalar returns only. 
    \item We provide a principled view of TD-ES as estimating the gradient of a smoothed objective and show (under standard smoothness) second-order accuracy in the smoothing scale; we explain how bounded support induces trust-region–like behavior without requiring backpropagation or KL constraints. 
    \item We specify a practical two-stage schedule (PPO pretraining $\rightarrow$ TD-ES refinement) with equal interaction budgets. 
    \item We demonstrate our approach improves PPO on a variety of robotic manipulation tasks, and demonstrate through ablation that this improvement is enhanced through our bounded-support approach.
\end{enumerate}

\section{Related Work} \label{sec:related-work}

Our approach builds on several complementary research directions in reinforcement learning and evolutionary optimization.
We review trust-region methods for policy optimization, gradient-evolution hybrid approaches, distribution design for ES, and recent work on triangular distributions in an RL context.

\subsubsection{Trust-Region Policy Optimization}

Trust-region methods such as TRPO~\cite{Schulman2015-re} and PPO~\cite{Schulman2017-qh} localize updates by constraining the policy’s change (via KL penalties or clipping), a design that has proven dependable in locomotion and manipulation \cite{Heess2017-nv,KUO2023102009}. 
Importantly, these algorithms approximate gradients from stochastic rollouts; near competent policies the signals become small-magnitude and high-variance, and performance can hinge on sensitive hyperparameters \cite{Engstrom2020-dk,Wolczyk2024-gr,Kirkpatrick2017-ln}. 
Trust-region ideas have also been adapted to ES (e.g., TRES~\cite{Liu2019-jx}), typically by constraining action space divergences between successive policies. 
In contrast, our approach enforces parameter space locality during population generation by sampling from a bounded-support distribution, yielding a simple, compute-light mechanism that curbs destabilizing updates without backpropagation.

\subsubsection{Gradient–Evolution Hybrids}

A growing line of work mixes gradient-based updates with evolutionary search. Examples include joint-update schemes such as CEM-RL~\cite{Pourchot2018-gl} and PPO–ES mixtures~\cite{Sigaud2022-oz}, as well as interleaved variants (e.g., EPO~\cite{Mustafaoglu2025-na}). 
While these hybrids can improve performance, meta-analyses note added scheduling complexity and potential drift \cite{Lin2024-gf,Bai2023-xp}. 
We adopt a simpler sequential hand-off tailored to refinement: PPO is used first to enter a high-reward basin, after which a localized, gradient-free ES performs constrained search within that basin. This avoids running two optimizers concurrently while retaining the complementary strengths of each.

\subsubsection{Distribution Design for ES}

ES is attractive for massive parallelism and robustness to sparse rewards~\cite{Salimans2017-nv,Such2017-ui,Ha2018-jy,Conti2017-rb,Huizinga2018-fk}, yet its sample efficiency in high-dimensional policies suffers when isotropic Gaussian perturbations disperse candidates far from the incumbent potentially reducing sample efficiency \cite{Majid2021-rd,Wong2024-ms,Pagliuca2019-di}.
CMA-ES can improve locality via learned covariance~\cite{Hansen2016-fe,Uchida2024-pt}, but its $O(d^2)$ cost limits applicability with deep networks~\cite{Nishida2018-wq}. 
An orthogonal approach is to shape the perturbation law itself to bias exploration toward local moves; we follow this direction using bounded-support perturbations (see “Triangular Distributions For Reinforcement Learning” below and Section~\ref{sec:methodology}).

\subsubsection{Triangular Distributions For Reinforcement Learning}

Petersen et al.\ \cite{Petersen2024-td} explored triangular distributions in RL and provided an implementation we use. 
We differ by employing the triangular law for parameter space ES in a refinement setting, emphasizing bounded support to localize exploration and stabilize updates (see Section~\ref{sec:methodology}).

\section{Preliminaries} \label{sec:prelim}

\subsection{Problem Setting and Notation}

We consider an episodic Markov decision process (MDP) with state space~$\mathcal{S}$, action space~$\mathcal{A}$, horizon $H$, and stochastic dynamics $p(s_{t+1}\mid s_t,a_t)$. 
At each time step the agent receives a scalar reward $r_t = r(s_t,a_t)$. 
Under a policy $\pi_\theta(a\mid s)$, parameterized by $\theta\in\mathbb{R}^{d}$, an episode generates a trajectory $\tau=(s_0,a_0,r_0,\dots,s_{H-1},a_{H-1},r_{H-1})$ according to $\pi_\theta$ and $p$.

We maximize the expected discounted return:
\begin{equation}
    J(\theta)=
    \mathbb E_{\tau\sim\pi_\theta}\Bigl[\sum_{t=0}^{H-1}\gamma^{t}r_t\Bigr],
    \qquad \gamma\in(0,1]
  \label{eq:return}
\end{equation}
Unless stated otherwise we use the diagonal-Gaussian policy
\begin{equation}
    \pi_\theta(a\mid s)=
    \mathcal N\!\bigl(
    a;\mu_\theta(s),\operatorname{diag}\,\sigma_\pi^{2}(s)\bigr),
    \label{eq:gauss-policy}
\end{equation}
where the network outputs the mean $\mu_\theta(s)$ and log-standard deviation $\log\sigma_\pi(s)$.
We distinguish two noise scales: $\sigma_\pi$ governs on-policy exploration in \emph{action space}, while a separate scale $\sigma_{\mathrm{ES}}$ will control \emph{parameter space} perturbations in the evolutionary refinement stage.

\subsection{Evolution Strategies Recap} 

In classic ES, at step $t$, a population of $n$ perturbed parameter vectors is drawn from an isotropic Gaussian centered at the current point $\theta_t$,
\begin{equation}
    \theta_t^{(i)} = \theta_t +
    \sigma_{\mathrm{ES}}\,\delta_t^{(i)}, 
    \qquad
    \delta_t^{(i)} \sim \mathcal N(0, I),
    \label{eq:es-sampling}
\end{equation}
where the superscript $(i)$ denotes the $i^{\text{th}}$ individual in the current population.

Each $\theta_i$ instantiates a candidate policy $\pi_{\theta_{i}}$.
After parallel rollouts we form the Monte Carlo score-function (likelihood-ratio) gradient estimator~\cite{Salimans2017-nv} and update the center as in \eqref{eq:es-update}.
\begin{equation}
    g_t=\frac{1}{n\sigma_{\mathrm{ES}}}\sum_{i=1}^{n}J(\theta_i)\,\delta_i,
    \qquad
    \theta_{t+1}=\theta_t+\alpha_t g_t.
    \label{eq:es-update}
\end{equation}
This formulation communicates only scalar returns, making ES trivially parallelizable. 

The gradient-free nature of ES, combined with its robustness to local optima~\cite{Salimans2017-nv} and embarrassingly parallel structure, makes it particularly well-suited for policy refinement scenarios where gradient-based methods may encounter diminishing returns near competent policies. 

However, isotropic noise explores the entire parameter space, which can be inefficient when the policy is already near a good solution. We make the connection between \eqref{eq:es-update} and a policy-gradient proxy precise in the following section.

\subsection{ES as a Policy-Gradient Proxy} \label{sec:es-gradient}

Classical policy-gradient methods estimate $\nabla_\theta J(\theta)$ for the objective in \eqref{eq:return} by injecting randomness in \emph{action space} under the policy \eqref{eq:gauss-policy}.
ES provides an alternative in \emph{parameter space}: sample perturbations $\delta$ as in \eqref{eq:es-sampling}, evaluate returns $J(\theta+\sigma_{\mathrm{ES}}\delta)$ for the same objective \eqref{eq:return}, and aggregate them into the Monte Carlo estimator \eqref{eq:es-update}.
Formally, define the Gaussian-smoothed objective
\begin{equation}
    J_\sigma(\theta)
    \;=\;
    \mathbb{E}_{\delta\sim\mathcal N(0,I)}
    \bigl[J(\theta+\sigma_{\mathrm{ES}}\delta)\bigr].
    \label{eq:smoothed-objective}
\end{equation}
The gradient of $J_\sigma(\theta)$ is
\begin{equation}
    \nabla_\theta J_\sigma(\theta)
    \;=\;
    \frac{1}{\sigma_{\mathrm{ES}}}
    \mathbb{E}_{\delta\sim\mathcal N(0,I)}
    \bigl[J(\theta+\sigma_{\mathrm{ES}}\delta)\,\delta\bigr].
    \label{eq:gradient}
\end{equation}
Consequently, the Monte-Carlo average $g_t$ in \eqref{eq:es-update} is an unbiased estimator of
the exact gradient in \eqref{eq:gradient}, yet it needs only scalar returns.

In other words, ES performs policy gradient by approximating $\nabla_\theta J$ with $\nabla_\theta J_\sigma$. Under mild smoothness ($J\in C^2$), a second-order Taylor expansion gives $\nabla_\theta J_\sigma(\theta)=\nabla_\theta J(\theta)+\mathcal O(\sigma_{\mathrm{ES}}^{2})$; thus smaller $\sigma_{\mathrm{ES}}$ lowers smoothing bias but raises Monte Carlo variance, mitigated by larger populations, antithetic pairs, and a constant baseline.

Because the estimator uses sampled perturbations, their \emph{distribution and scale} determine variance and the locality of search; in refinement, unconstrained isotropic exploration is often wasteful. As discussed in Section~\ref{sec:related-work}, on-policy methods such as PPO also approximate gradients from rollouts but become small-magnitude and high-variance near competent policies. Consequently, choosing the perturbation distribution is crucial for efficient late-stage improvement; in Section~\ref{sec:methodology} we instantiate this with a bounded-support sampler that concentrates exploration near $\theta$ and stabilizes refinement.

\subsection{Triangular Distribution} \label{sec:td-prelim}

We consider the symmetric triangular distribution with mode at zero and support $[-a,a]$, whose probability density function is
\begin{equation}
    f(x) = \begin{cases}
        \frac{1}{a}\left(1 - \frac{|x|}{a}\right), & \text{if } |x| \leq a, \\
        0, & \text{otherwise}.
    \end{cases}
\end{equation}
Here $a>0$ is the \emph{half-width} (the distance from the mode to either boundary). For this distribution, the mean is $0$ and the variance is $\operatorname{Var}(X)=a^{2}/6$.

In our implementation, we set the half-width $a=\sigma_\mathrm{ES}$.
This yields perturbations $\varepsilon$ with support $[-\sigma_\mathrm{ES},\sigma_\mathrm{ES}]$ and can be sampled as $\varepsilon=\sigma_\mathrm{ES}(U-V)$ where $U,V \sim \mathrm{Uniform}(0,1)$ and are independent.
This parameterization directly controls the maximum parameter exploration, and enables a natural comparison between the bounded-support triangular distribution and unbounded Gaussian perturbations at the same scale parameter.

\begin{figure}[h]
\centering
\includegraphics[width=0.8\columnwidth]{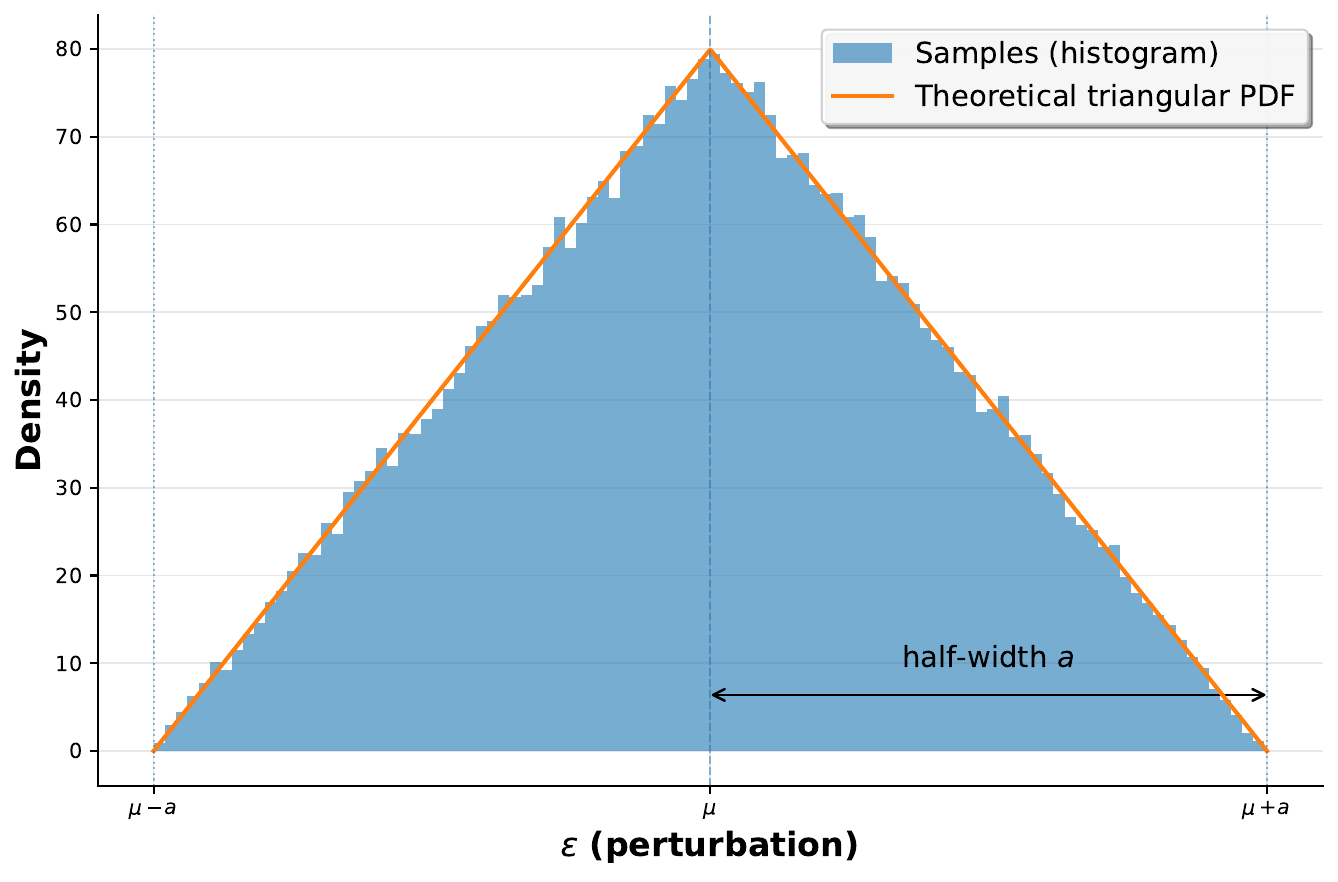}
\caption{Theoretical triangular distribution PDF overlaid with a subset of our actual samples (histogram).}
\label{fig:triangle-pdf}
\end{figure}

Figure~\ref{fig:triangle-pdf} shows the theoretical triangular PDF with support $[\mu-a,\mu+a]$ (here $\mu=0$) with the half-width $a$ indicated. 
We overlay this with a subset of our samples from an experiment to verify that the sampler matches the theoretical distribution.

\section{Methodology} \label{sec:methodology}

\subsection{Problem Formulation} \label{sec:formulation}

We operate in a policy refinement context, where the initial anchor policy $\pi_{\theta_0}$, parameterized by a neural network, is obtained through training with PPO~\cite{Schulman2017-qh}, chosen for its empirical reliability on robotic tasks \cite{Mittal2023-uf,Heess2017-nv,KUO2023102009,Rupam-Mahmood2018-nu}.

This anchor policy is safe and reasonably successful, but not yet optimal. 
Our objective is to find a refined parameter vector $\theta^\star$ where

\begin{equation}
    \theta^\star \;=\;\arg\max_{\theta}\;J(\theta).
    \label{eq:pf}
\end{equation}
Although we consider a policy refinement context when presenting our methods,
the full benefits of our work can be harnessed by invoking a 2-stage sequential
training framework. This is discussed further in Section~\ref{sec:sequential}.

\subsection{Bounded-Support Perturbations for ES} \label{sec:bounded-support}

We refine the PPO anchor $\theta_0$ using Evolution Strategies. At iteration $t$ we maintain a center $\theta_t$ (initialized as $\theta_0$) and replace the isotropic Gaussian sampler \eqref{eq:es-sampling} with a bounded-support, zero-mean, factorized base distribution $q$ to localize search and improve refinement efficiency.

\begin{figure}[ht]
  \centering
  \includegraphics[width=0.7\columnwidth]{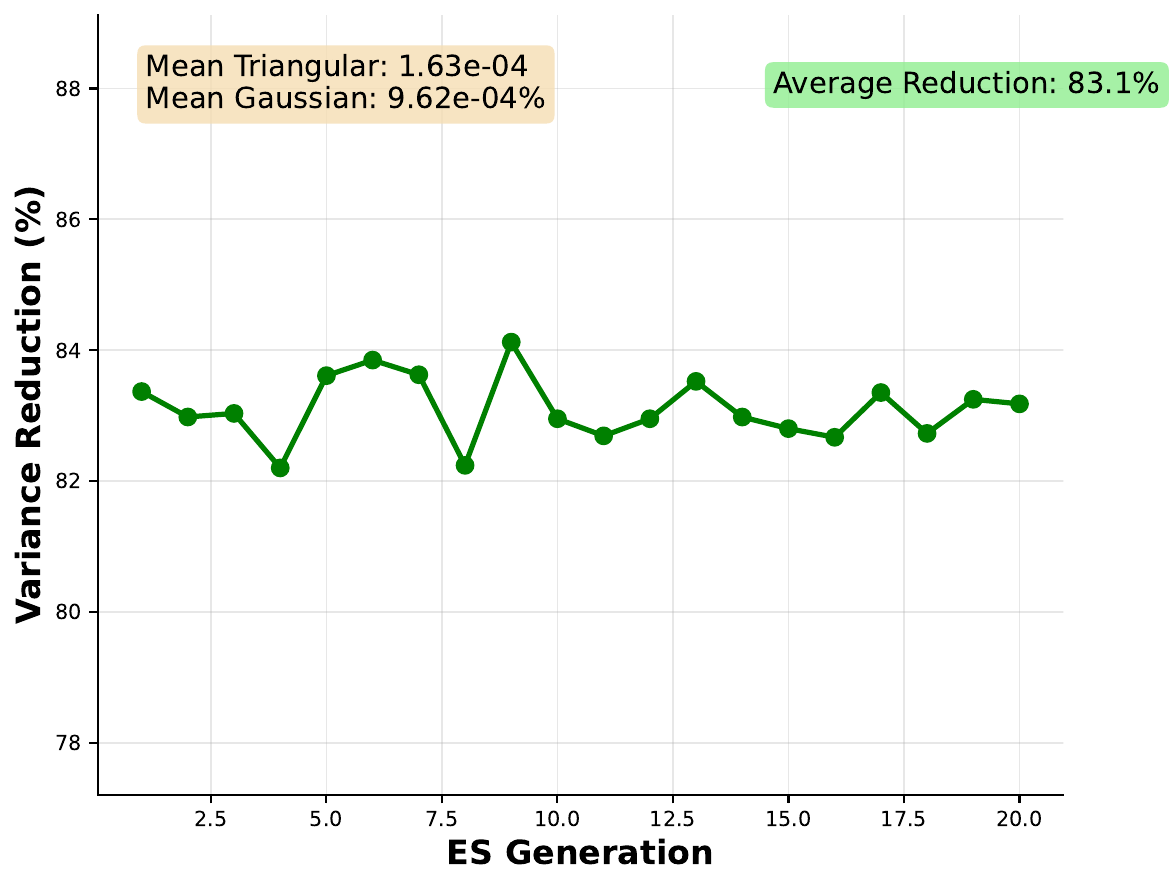}
    \caption{Relative reduction in gradient estimator variance achieved by triangular perturbations compared to Gaussian perturbations during ES refinement. The y-axis shows the percentage by which triangular ES reduces variance relative to Gaussian ES at each generation, computed from multiple independent gradient estimates. Positive values indicate lower variance for triangular perturbations. The bounded support of triangular distributions consistently reduces estimator variance throughout the refinement process, demonstrating the stabilizing effect of localized parameter-space exploration.}
\label{fig:gradient-variance}
\end{figure}

Concretely, we use the symmetric triangular law on each coordinate with support $[-\sigma_\mathrm{ES},\sigma_\mathrm{ES}]$ and mode $0$ (detailed in \ref{sec:td-prelim}),
\begin{equation}
    \begin{aligned}
        q_\triangle(\varepsilon_k) &=
        \begin{cases}
        \frac{1}{\sigma_{\mathrm{ES}}}\left(1 - \frac{|\varepsilon_k|}{\sigma_{\mathrm{ES}}}\right), 
        & \text{if} |\varepsilon_k| \le \sigma_{\mathrm{ES}}, \\
        0, & \text{otherwise},
        \end{cases} \\
        q(\varepsilon) &= \prod_{k=1}^d q_\triangle(\varepsilon_k).
    \end{aligned}
    \label{eq:tri-pdf}
\end{equation}
Given $\varepsilon_i\!\sim\!q$ we construct $n=2m$ candidates via antithetic pairs
\begin{equation}
    \theta^{\pm}_i \;=\; \theta_t \,\pm\, \sigma_{\mathrm{ES}}\,\varepsilon_i,
    \qquad i=1,\dots,m.
    \label{eq:tdes-antithetic}
\end{equation}
The bounded support implies $\|\theta_i^{\pm}-\theta_t\|_\infty \le \sigma_{\mathrm{ES}}$ while the mode at zero concentrates probability on small perturbations and further reduces estimator variance by concentrating sampling probability on perturbations near the current parameters, where the objective function is more likely to exhibit locally smooth behavior. 
Figure~\ref{fig:gradient-variance} empirically demonstrates this effect, showing that triangular perturbations achieve 83.1\% average variance reduction compared to Gaussian perturbations during refinement.

\subsection{Gradient Approximation under Bounded-Support ES} \label{sec:tdes-gradient}

Given antithetic pairs $\{\theta_i^{\pm}\}_{i=1}^{m}$ from \eqref{eq:tdes-antithetic}, let $J_i^{\pm}=J(\theta_i^{\pm})$ be their scalar returns and set $n=2m$. 
We compute centered ranks within the generation by ranking the $n$ returns from worst to best, mapping ranks to $[-\tfrac{1}{2},\,\tfrac{1}{2}]$, and standardizing to unit variance; denote the resulting scores by $\tilde J_i^{\pm}$. 
Our update forms a finite-difference search direction using these scores:
\begin{equation}
    \begin{aligned}
    & g_t = \frac{1}{m\,\sigma_{\mathrm{ES}}}\sum_{i=1}^{m}\bigl(\tilde J_i^{+}-\tilde J_i^{-}\bigr)\,\varepsilon_i,\\
    & \theta_{t+1} = \theta_t + \alpha_t\, g_t .
    \label{eq:tdes-estimator}
    \end{aligned}
\end{equation}
Antithetic pairing cancels even-order terms in the finite difference and reduces variance. With standardized, zero-mean perturbations ($\mathbb{E}[\varepsilon]=0$, $\mathbb{E}[\varepsilon\varepsilon^\top]=I$) and smooth $J$, \eqref{eq:tdes-estimator} is a \emph{constant-scaled central-difference proxy}: when raw returns are used its expectation is
\begin{equation}
      \mathbb{E}[g_t] \;=\; c\,\nabla_\theta J(\theta_t) \;+\; \mathcal{O}(\sigma_{\mathrm{ES}}^{2}),
\end{equation}
for some constant $c>0$. In our implementation with a prefactor $1/(m\sigma_{\mathrm{ES}})$ we have $c=2$, while the normalized two-sided estimator $\tfrac{1}{2m\sigma_{\mathrm{ES}}}\sum_{i=1}^m (J_i^+-J_i^-)\varepsilon_i$ gives $c=1$. 
With centered ranks as our default, the estimator is scale- and shift-invariant, remains directionally aligned in the small-$\sigma_{\mathrm{ES}}$ regime, and exhibits the same $\mathcal{O}(\sigma_{\mathrm{ES}}^{2})$ truncation order; the effective constant depends on the score transform and local return distribution and is absorbed into $\alpha_t$.

\subsection{Trust-Region–Like Locality from Bounded Support} \label{sec:trust-region}

The sampler in \eqref{eq:tdes-antithetic} induces trust-region–like behavior directly in parameter space. 
First, bounded support guarantees a hard per-parameter radius (proportional to $\sigma_{\mathrm{ES}}$), capping the maximum excursion of any coordinate in a generation. 
Second, the triangular mode at zero concentrates probability on small perturbations, increasing the frequency of locally linear samples and reducing estimator variance. 
Empirically, these two effects (i) mitigate catastrophic regressions during refinement, (ii) improve the stability of late-stage updates without additional backpropagation or KL constraints, and (iii) preserve ES’s embarrassingly parallel rollout structure.

\subsection{Triangular-Distribution Evolution Strategies (TD-ES)}
\label{sec:tdes-alg}

We now summarize the full procedure as Triangular-Distribution Evolution Strategies (TD-ES) which combines the bounded-support sampler \eqref{eq:tdes-antithetic} with the centered-rank estimator \eqref{eq:tdes-estimator}.

\begin{algorithm}[ht]
\footnotesize
\caption{TD-ES}
\label{alg:tdes}
\DontPrintSemicolon
\SetKwInOut{Input}{Input}\SetKwInOut{Output}{Output}
\SetKwFunction{Rollout}{Rollout}
\SetKwProg{Fn}{Function}{}{}
\Input{anchor policy parameters $\theta_0$, pairs $m$, iterations $T$, perturb.\ scale $\sigma_{\mathrm{ES}}$, step size $\alpha$, action std $\sigma_a$,  decays $\lambda_\sigma$}
\Output{Refined parameters $\theta$}
$\theta \gets \theta_0$\;
\For{$t=1$ \KwTo $T$}{
    \For{$i=1$ \KwTo $m$}{
        sample $\varepsilon_i \sim q_\triangle$\
        \tcp*{triangular noise on $[-\sigma_\mathrm{ES},\sigma_\mathrm{ES}]$}
        
        $\theta_i^{\pm} \gets \theta \pm \sigma_{\mathrm{ES}}\,\varepsilon_i$\ \tcp*{antithetic candidates}

        $J_i^\pm \gets \textsc{Rollout}(\theta_i^\pm;\,\sigma_a)$ \tcp*{scalar episodic returns}
        
    }
    
    form centered ranks $\tilde J_i^\pm$ over $\{J_j^\pm\}$ \tcp*{unit variance}
    
    $g \gets \frac{1}{m\,\sigma_{\mathrm{ES}}}\sum_{i=1}^m (\tilde J_i^+ - \tilde J_i^-)\,\varepsilon_i$ \tcp*{cf. Eq.~\eqref{eq:tdes-estimator}}
    
    $\theta \gets \theta + \alpha\, g$ \tcp*{parameter update}

     $\sigma_{\mathrm{ES}} \gets \lambda_\sigma\,\sigma_{\mathrm{ES}}$
}
\Return{$\theta$}\;

\BlankLine
\Fn{\Rollout{$\theta;\,\sigma_a$}}{
  run one episode with $a_t \sim \mathcal N(\mu_\theta(s_t),\,\sigma_a^2 I)$\;
  \Return $\sum_{t=0}^{H-1}\gamma^t r_t$ \tcp*{$\gamma$ is the RL discount (not a decay)}
}

\end{algorithm}

\emph{Rollout notation.} As in Sec.~\ref{sec:prelim}, $s_t$ is the state at time $t$, $a_t$ is the sampled action, $r_t$ is the reward, $H$ is the horizon, and $\gamma$ is the RL discount; $\mu_\theta(\cdot)$ is the policy mean and $\sigma_a$ is the fixed action std used during ES (see~\ref{sec:imp-details}). \textsc{Rollout} returns one Monte Carlo sample of Eq.~\eqref{eq:return}.

\subsection{Two-Stage PPO and TD-ES Schedule} \label{sec:sequential}

We adopt a sequential hand-off approach: PPO for initial skill acquisition, followed by TD-ES for stable refinement.
Rather than using fixed budget allocations, we empirically determine the handoff point based on when PPO achieves a competent policy --- typically identified when learning curves show diminishing returns or when success rates reach a reasonable baseline (e.g., 40-60\% task completion).
This ensures TD-ES begins refinement from a meaningful anchor point rather than from random initialization. The total interaction budget remains fixed across all compared methods to ensure fair evaluation.

\section{Experimental Setup} \label{sec:experiment-setup}

Our experiments are designed to evaluate whether refinement with our bounded-support approach is able to benefit performance across a variety of robotic manipulation tasks. 
We also aim to identify how robust these benefits are across conditions. 
All experiments were run on a single NVIDIA RTX~4070 using Isaac Lab~\cite{Mittal2023-uf} as the GPU-accelerated physics simulator for robotic manipulation, via its vectorized environment interface for parallel rollouts.
For statistical robustness, we repeat every configuration with nine independent random seeds and reuse the same seed set across all methods. 
Fairness across methods is ensured as methods share identical observation/action spaces and identical total environment step budgets; specific budget allocations are detailed in Section~\ref{sec:imp-details}. 

We report success rates rather than cumulative reward because binary task completion is more interpretable and directly relevant to real-world deployment than engineered reward signals. 
Following best practices for reliable RL evaluation~\cite{Agarwal2021-hx}, aggregate performance is summarized with the Interquartile Mean (IQM) and 95\% stratified bootstrap confidence intervals, which yields robust estimates with fewer runs and reduces sensitivity to outliers. 
e additionally report the probability of improvement $\mathrm{P}(\text{TD-ES} \rightarrow \text{PPO})$ using the Mann–Whitney U statistic to quantify the likelihood that our method outperforms the baseline. 
For individual task analysis, per-task results are presented as mean~$\pm$~standard deviation. 

Section~\ref{sec:task-domains} outlines the tasks/environments used for evaluation, and Section~\ref{sec:imp-details} lists specific implementation details to support reproducibility. 
Section~\ref{sec:results}  then proceeds in three parts: (i) Aggregate Performance versus existing baseline to establish gains, (ii) a Per-Task Analysis to localize where improvements concentrate, and (iii) a Performance Distribution Analysis to examine robustness across difficulty factors and tail behavior (5th/95th percentiles).

\subsection{Task Domains} \label{sec:task-domains}

\begin{figure}[ht]
  \centering
  \begin{minipage}[t]{0.32\columnwidth}
    \centering
    \includegraphics[height=0.75in,width=1.0in]{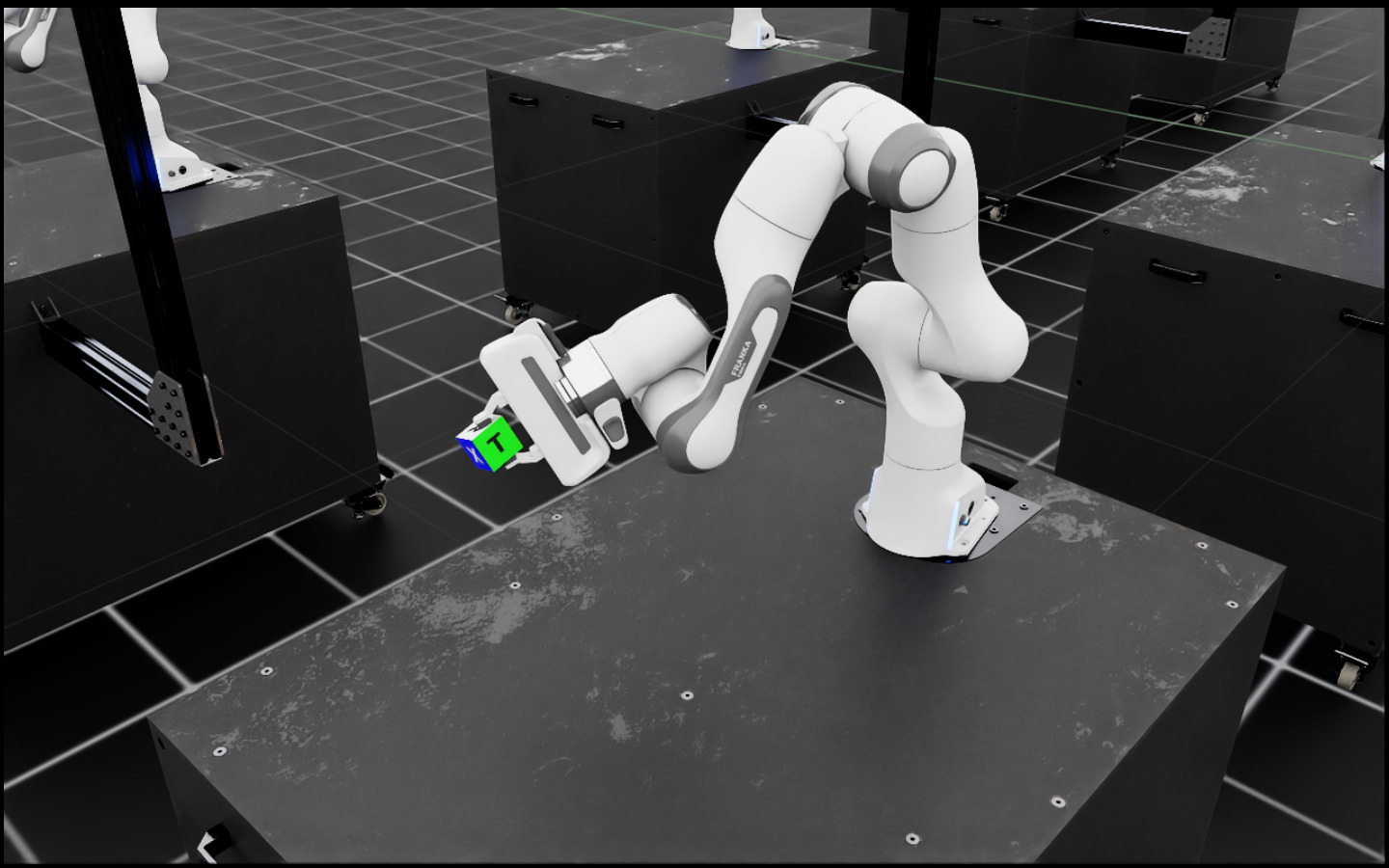}
    \vspace{2pt}\par\footnotesize (a)
  \end{minipage}\hfill
  \begin{minipage}[t]{0.32\columnwidth}
    \centering
    \includegraphics[height=0.75in,width=1.0in]{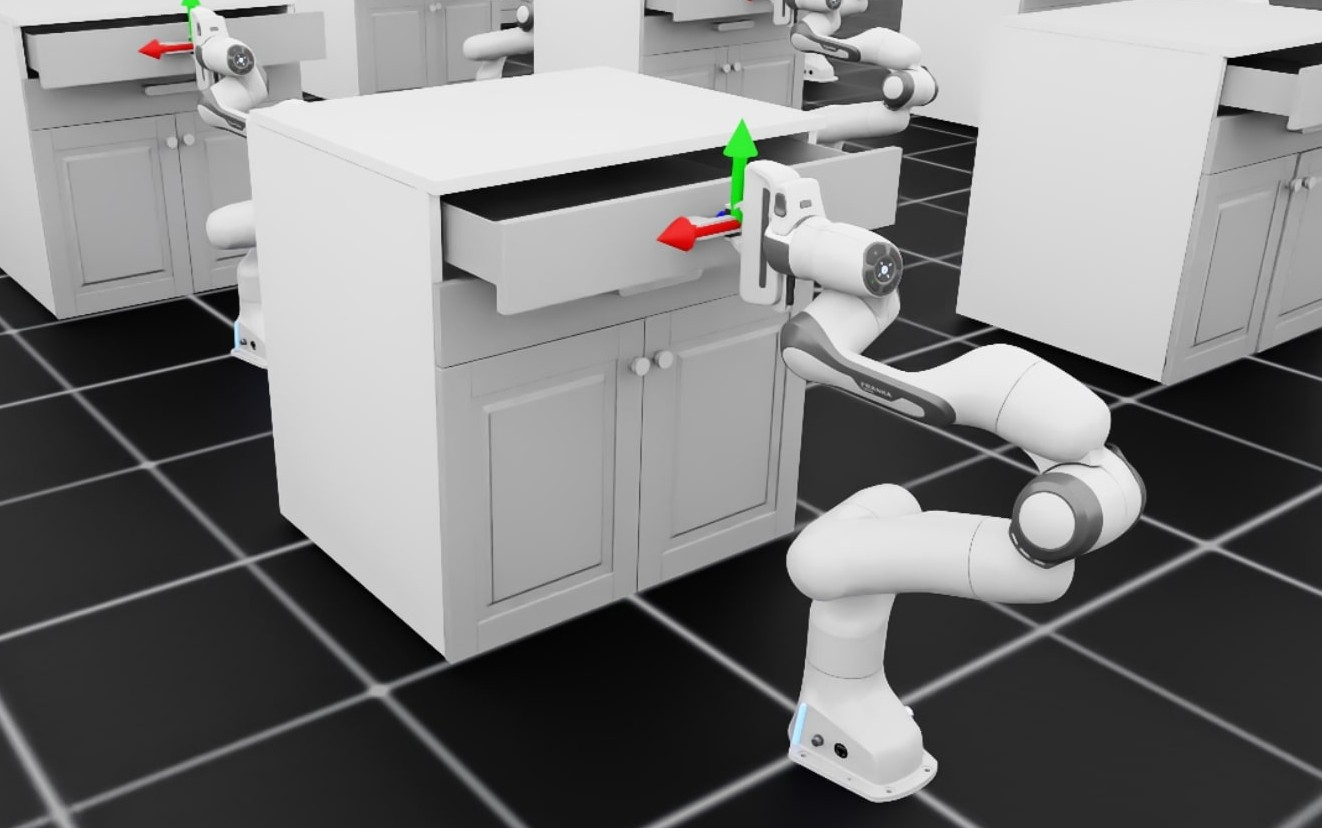}
    \vspace{2pt}\par\footnotesize (b)
  \end{minipage}\hfill
  \begin{minipage}[t]{0.32\columnwidth}
    \centering
    \includegraphics[height=0.75in,width=1.0in]{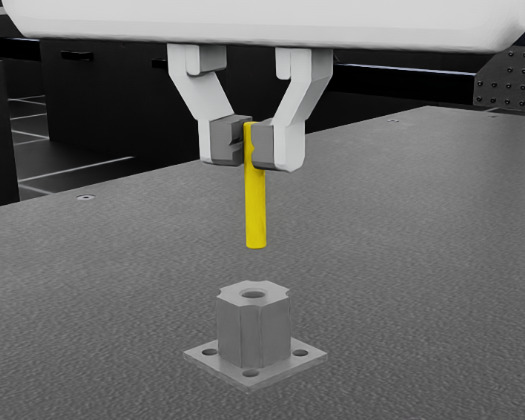}
    \vspace{2pt}\par\footnotesize (c)
  \end{minipage}
  \caption{Robotic manipulation tasks: (a) Lift-Cube with 36D observations and 4096 environments, (b) Open-Drawer with 31D observations and 4096 environments, (c) Peg-Insert with 19D observations and 512 environments due to contact modeling demands.}
  \label{fig:tasks}
\end{figure}

We evaluate our approach on three robotic manipulation tasks using a 7-DoF Franka Panda arm, chosen to test policy refinement across different precision requirements. Lift-Cube and Open-Drawer use 8D joint-space control (7 joints + gripper), while Peg-Insert employs 6D task-space control with constrained roll/pitch for tool alignment. 
Figure~\ref{fig:tasks} provides visual overview and technical specifications.

\textbf{Lift-Cube} requires grasping and positioning a cube within 4cm of a target. Refinement challenges include achieving spatial precision while maintaining grasp stability, where small parameter improvements reduce tolerance violations and grip failures.

\textbf{Open-Drawer} involves grasping a handle and opening a drawer at least 30cm while maintaining handle contact. Success depends on reliable initial grasping, coordinated pulling motion, and sufficient extension distance --- all areas where fine-grained parameter adjustments provide significant benefits.

\textbf{Peg-Insert} demands inserting an 8mm peg into an 8mm hole (0.114mm clearance) with sub-millimeter positioning accuracy. Success requires the peg within 2.5mm laterally and 1mm above target depth. This contact-rich task exemplifies high-precision manipulation where minor control improvements dramatically impact insertion success by reducing jamming and misalignment.

\subsection{Implementation Details}
\label{sec:imp-details}

We use task-specific budgets with PPO$\rightarrow$TD-ES splits based on learning dynamics: Lift-Cube (12,000 steps, 83:17 split), Open-Drawer (19,200 steps, 67:33), Peg-Insert (4,470 steps, 67:33). 
PPO uses standard hyperparameters ($\gamma=0.99$, $\lambda=0.95$, $\epsilon=0.2$) with task-specific learning rates and network architectures. 
TD-ES initializes from PPO checkpoints with $\sigma_a=0.01$, decay $\lambda_\sigma=0.99$, and floor $\sigma_{\min}=1\times10^{-3}$. Key hyperparameters are shown in Table~\ref{tab:hyperparameters}.

\begin{table}[ht]
  \caption{Key task-specific hyperparameters.}
  \label{tab:hyperparameters}
  \centering
  \small
  \setlength{\tabcolsep}{4pt}
  \renewcommand{\arraystretch}{1.1}
  \begin{tabularx}{\columnwidth}{l*{3}{C}}
    \hline
    \bfseries Parameter & \bfseries Lift-Cube & \bfseries Open-Drawer & \bfseries Peg-Insert \\
    \hline
    PPO learning rate        & $1\times10^{-4}$ & $5\times10^{-4}$ & $1\times10^{-3}$ \\
    PPO Episodes per update      & 24               & 96               & 24 \\
    TD-ES $\sigma_{\text{ES}}$ & 0.0125         & 0.025            & 0.03 \\
    TD-ES learning rate      & 0.005            & 0.01             & 0.01 \\
    \hline
  \end{tabularx}
\end{table}

We compare our method against PPO-only (full budget without refinement) and Gaussian ES (identical to TD-ES but with unbounded Gaussian perturbations), isolating the bounded-support contribution.

\section{Results} \label{sec:results}

\begin{table*}[ht]
  \caption{Aggregate success rate across all tasks using robust statistical measures.}
  \label{tab:main-results}
  \centering
  \setlength{\tabcolsep}{4pt}
  \renewcommand{\arraystretch}{1.1}
  {\small
  \begin{tabularx}{\textwidth}{l*{3}{C}}
    \hline
    \bfseries Method & \bfseries IQM (95\% CI) & \bfseries Mean (95\% CI) & \bfseries P(Improvement vs PPO) \\
    \hline
    PPO & 67.2\% (55.8--77.5\%) & 65.3\% (56.0--74.6\%) & -- \\
    PPO$\rightarrow$Gaussian ES & 73.7\% (62.4--83.3\%) & 69.9\% (60.6--79.1\%) & 55.6\% (39.6-70.7)\% \\
    PPO$\rightarrow$TD-ES & \textbf{85.0\% (77.7-91.9\%)} & \textbf{83.6\% (78.2--88.9\%)} & \textbf{73.7\% (59.4--86.9\%)} \\
    \hline
  \end{tabularx}
  }
\end{table*}

\begin{table*}[ht]
  \caption{Per-task success rates (mean $\pm$ std) across experimental runs.}
  \label{tab:per-task-results}
  \centering
  \setlength{\tabcolsep}{4pt}
  \renewcommand{\arraystretch}{1.1}
  {\small
  \begin{tabularx}{\textwidth}{l*{4}{C}}
    \hline
    \bfseries Task & \bfseries PPO & \bfseries PPO$\rightarrow$Gaussian ES & \bfseries PPO$\rightarrow$TD-ES \\
    \hline
    Lift-Cube  & \mbox{75.2\% $\pm$ 18.8\%} & \mbox{76.0\% $\pm$ 19.6\%} & \bfseries\mbox{80.6\% $\pm$ 15.2\%} \\
    Open-Drawer& \mbox{67.0\% $\pm$ 31.4\%} & \mbox{75.8\% $\pm$ 33.2\%} & \bfseries\mbox{97.8\% $\pm$ 1.9\%}  \\
    Peg-Insert & \mbox{53.8\% $\pm$ 19.4\%} & \mbox{57.8\% $\pm$ 15.1\%} & \bfseries\mbox{72.3\% $\pm$ 5.7\%}  \\
    \hline
  \end{tabularx}
  }
\end{table*}

\subsection{Aggregate Performance}

Table~\ref{tab:main-results} presents aggregate task success rates across all methods. Our results demonstrate that ES-based refinement provides consistent improvements over PPO alone, with TD-ES offering additional systematic benefits.

The results show a clear progression in refinement effectiveness. TD-ES achieves an IQM success rate of 85.0\%, compared to Gaussian ES's 73.7\% and PPO's 67.2\%.
This demonstrates systematic improvements through ES-based refinement, with our bounded-support approach providing additional gains beyond standard Gaussian perturbations. 
The Mann-Whitney U test indicates 73.7\% probability that TD-ES outperforms PPO. We also found that there was a \textbf{69.2\% probability that TD-ES outperforms Gaussian ES}, demonstrating statistically significant improvement across the method progression.

\subsection{Per-Task Analysis}

Table~\ref{tab:per-task-results} reveals how refinement benefits scale with task requirements.

The results demonstrate systematic improvements through the refinement progression.
Open-Drawer shows the most dramatic benefits from bounded-support refinement, with TD-ES achieving near-perfect success (97.8\%) while drastically reducing variance ($\sigma$: 31.4\% → 1.9\%). 
On the Peg-Insert task our approach shows a substantial 18.5 percentage point improvement over the PPO baseline, with a much tighter clustering. Lift-Cube shows a consistent 5.4 percentage point improvement with reduced variance.

The variance reduction pattern is particularly striking: TD-ES consistently achieves the lowest performance variability across all tasks, with the most pronounced stabilization occurring on precision-demanding tasks. 
This suggests that bounded-support perturbations effectively prevent destabilizing parameter excursions during refinement.

\subsection{Bounded-Support Ablation}

To isolate the bounded-support contribution, we directly compare PPO$\rightarrow$TD-ES against PPO$\rightarrow$Gaussian ES using identical refinement protocols. 
Both methods initialize from the same PPO checkpoints and use identical hyperparameters, with the only difference being triangular versus Gaussian perturbation distributions.

The comparison reveals systematic advantages of bounded-support perturbations. 
At the aggregate level, TD-ES achieves 85.0\% IQM success compared to Gaussian ES's 73.7\%, representing a 15\% relative improvement in refinement effectiveness. 
A Mann-Whitney U test indicated 73.7\% probability that TD-ES outperforms PPO compared to 55.6\% for Gaussian ES, demonstrating stronger statistical evidence for TD-ES improvements. 
Further, another Mann-Whitney U test indicated that there is a 69.2\% chance that TD-ES actually outperforms Gaussian ES, further highlighting the benefits of the bounded-support approach.

The per-task analysis reveals task-dependent benefits of bounded-support perturbations. 
For Lift-Cube, both methods provide similar modest improvements over PPO, with TD-ES achieving slightly higher success (80.6\% vs 76.0\%) and reduced variance (15.2\% vs 19.6\%). 
Open-Drawer shows the most dramatic bounded-support advantage: while Gaussian ES provides moderate improvement (67.0\% → 75.8\%), TD-ES achieves near-perfect performance (97.8\%) with drastically reduced variance (1.9\% vs 33.2\%). 
Peg-Insert demonstrates consistent bounded-support benefits, with TD-ES substantially outperforming Gaussian ES (72.3\% vs 57.8\%) while maintaining much tighter performance clustering.

The variance reduction pattern is particularly revealing: TD-ES consistently achieves lower performance variability than Gaussian ES across all tasks, with the effect being most pronounced on precision-demanding tasks. 
This suggests that bounded-support perturbations provide enhanced stability during refinement by preventing parameter excursions that degrade performance in high-precision scenarios.

Importantly, hyperparameters were tuned for triangular distributions and may not be optimal for Gaussian perturbations. However, this controlled comparison demonstrates that bounded-support perturbations systematically enhance ES refinement effectiveness beyond what standard Gaussian perturbations can achieve, providing strong evidence for our bounded-support hypothesis in policy refinement contexts.

\begin{figure}[t!]
\centering
\includegraphics[width=\linewidth]{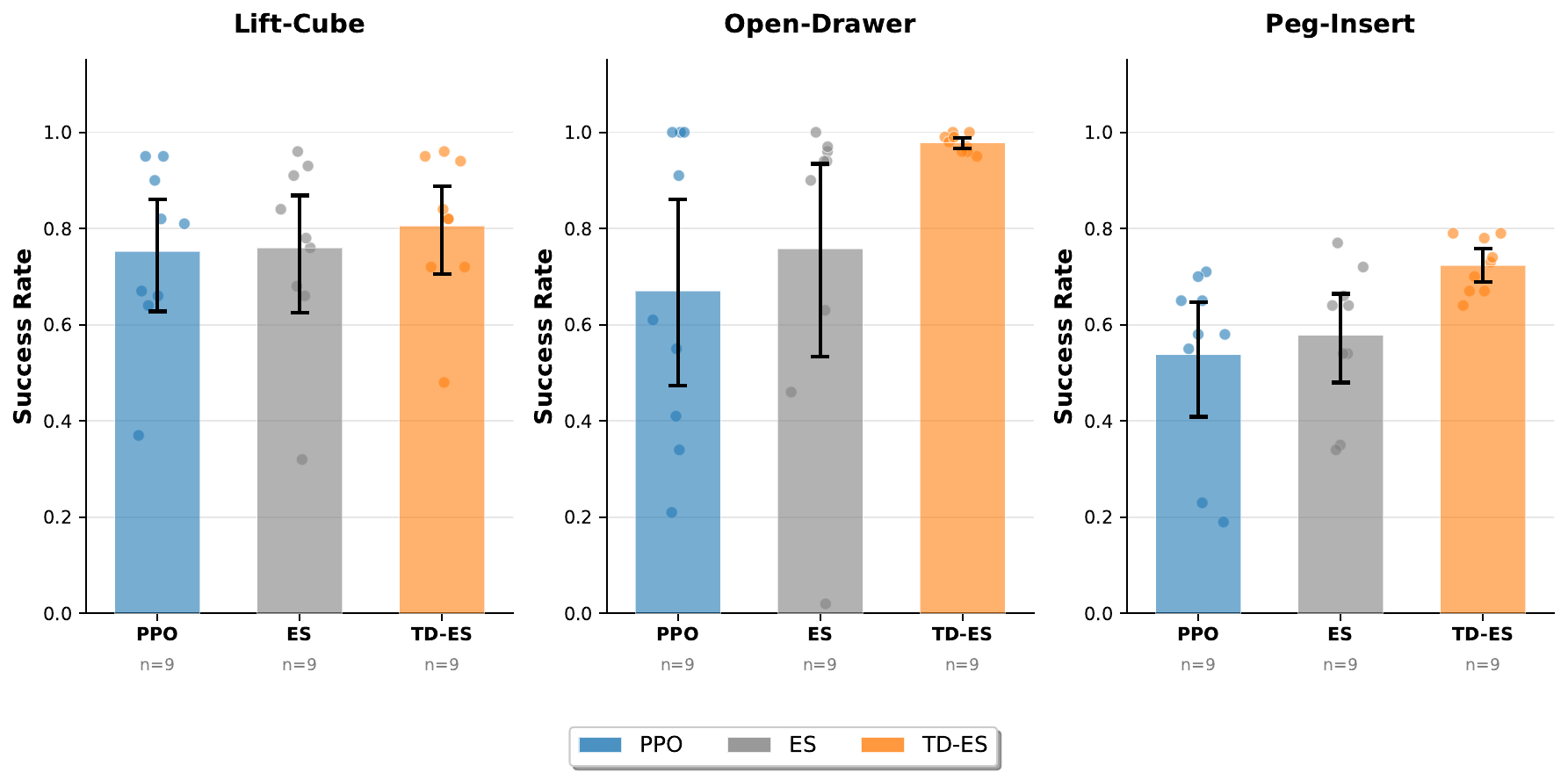}
\caption{Individual run success rates. Our approach shows reduced variance across all tasks, with particularly tight clustering on precision-demanding tasks (Open-Drawer, Peg-Insert).}
\label{fig:per-task}
\end{figure}

\subsection{Performance Distribution Analysis}

Figure~\ref{fig:per-task} shows the distribution of individual run performance across each task.

The individual run distributions clearly illustrate the systematic benefits of our bounded-support refinement approach. 
The three-method comparison shows consistent ranking: PPO provides the baseline, Gaussian ES offers moderate improvements with increased variance in some cases, and TD-ES delivers the strongest performance with reduced variability.

\begin{figure}[!t]
\centering
\includegraphics[width=\linewidth]{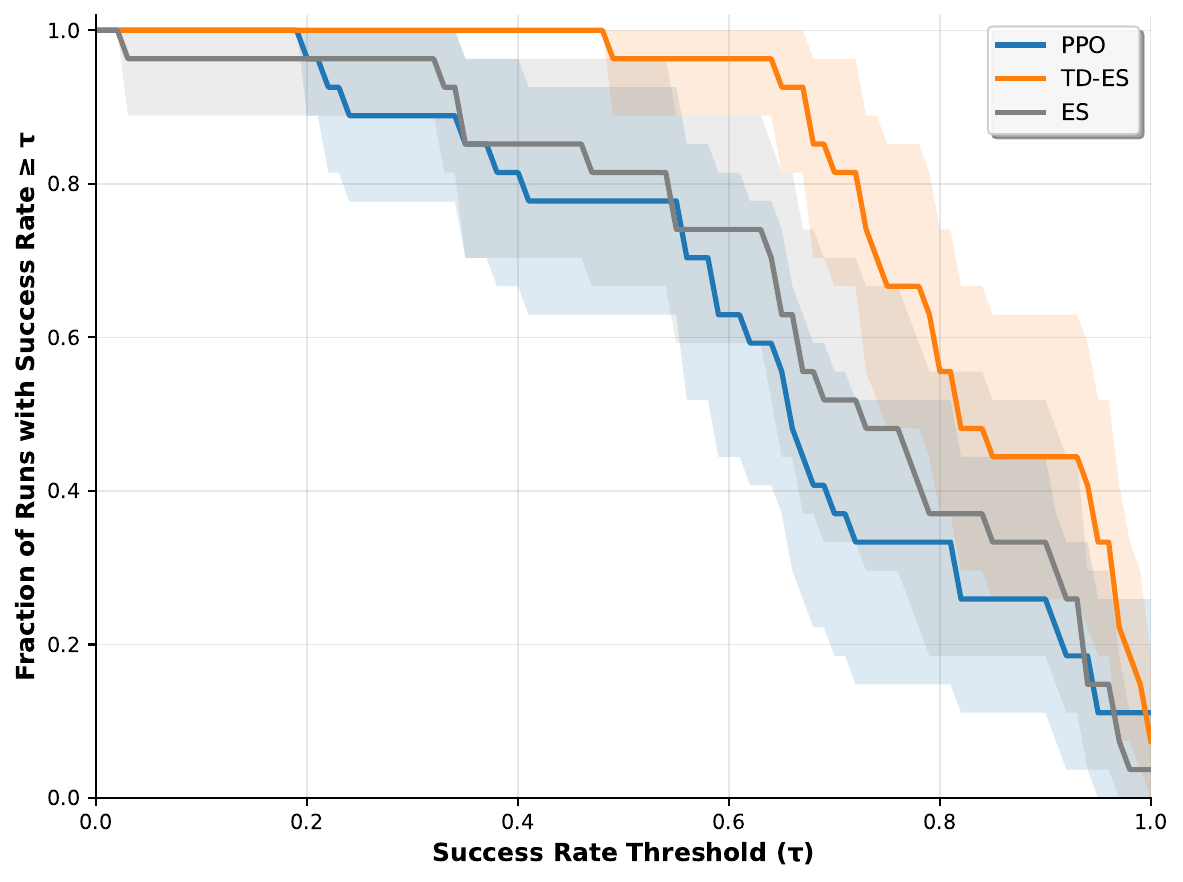}
\caption{Performance profiles showing fraction of runs achieving success rates above threshold $\tau$. TD-ES curve dominates PPO across all thresholds, indicating more reliable high-performance outcomes.}
\label{fig:performance-profiles}
\end{figure}

As recommended by Agarwal et al. \shortcite{Agarwal2021-hx}, we plot performance profiles in Figure~\ref{fig:performance-profiles} to demonstrate the cumulative distribution of success rates across all experimental runs. 

This performance profile demonstrates that our approach is superior across all success rate thresholds. The TD-ES curve remains consistently above PPO, with particularly pronounced advantages in the high-performance region ($\tau$ > 0.7). At a 80\% success threshold, TD-ES achieves this performance in approximately 50\% of runs compared to PPO's 20\%, highlighting the practical reliability improvements for deployment scenarios where consistent task completion is critical.

\section{Conclusion and Future Work}

We have presented Triangular-Distribution Evolution Strategies (TD-ES), a bounded-support approach for policy refinement that addresses gradient-based limitations in late-stage training. Our two-stage PPO$\rightarrow$TD-ES framework demonstrates systematic improvements over both PPO alone and standard Gaussian ES across robotic manipulation tasks, with particularly pronounced benefits on precision-demanding scenarios.

The key insight is that bounded-support perturbations provide trust-region-like behavior in parameter space without backpropagation or KL constraints. By concentrating exploration near competent policies while preventing destabilizing excursions, TD-ES achieves improved performance and enhanced reliability. Results show clear progression: PPO establishes baselines, Gaussian ES provides moderate benefits, and TD-ES delivers substantial improvements with reduced variance.

Several limitations are deserve mention. Refinement effectiveness remains constrained by anchor policy quality, with performance varying across seeds. Computational constraints limited evaluation to moderate-complexity tasks, leaving scaling to more demanding scenarios an open question.

Future work offers promising directions. Stronger pretraining methods may reduce seed sensitivity and establish higher-quality anchors. Alternative bounded-support distributions --- including Beta, Kumaraswamy, trapezoidal, raised-cosine, and truncated Gaussian forms --- may better capture actuator constraints and provide smoother gradient approximations. Finally, extending evaluation to complex manipulation suites would strengthen understanding of when these methods excel.

The systematic improvements demonstrated suggest that bounded-support perturbations offer a principled, practical approach to policy refinement with broader potential in robotics and beyond.

\bibliography{references}
\bibliographystyle{named}

\end{document}